\documentclass[letterpaper, 10 pt, conference]{ieeeconf}  

\IEEEoverridecommandlockouts                              

\usepackage{graphics} 
\usepackage{epsfig} 
\usepackage{mathptmx} 
\usepackage{times} 
\usepackage{amsmath} 
\usepackage{dsfont}
\usepackage{amssymb}  
\usepackage{float}
\usepackage{booktabs}
\usepackage{tabularx} 
\usepackage{makecell} 
\usepackage{url}
\usepackage{microtype}

\title{\LARGE \bf
Affective Shared Autonomy: Temporal Affect Dynamics and Subjective Evaluation in Bimanual Teleoperation Tasks
\vspace{-0.14in}
}

\author{
Zhengji Liang$^{1}$, Guiyin Tian$^{1}$, Sijin Qu$^{2}$, Hainan Liu$^{1}$, and Shiyan Hu$^{1}$%
\thanks{$^{1}$Department of Data and Systems Engineering, The University of Hong Kong, Hong Kong SAR, China.}%
\thanks{$^{2}$Department of Electrical and Computer Engineering, The University of Hong Kong, Hong Kong SAR, China.}%
}

\begin{document}

\maketitle
\thispagestyle{empty}
\pagestyle{empty}

\begin{abstract}
Physical teleoperation integrates human cognitive flexibility with robotic precision, yet demanding manipulation tasks frequently induce severe cognitive workload, acute frustration, and execution breakdown. Conventional shared autonomy paradigms rely primarily on task-based rules, such as spatial error boundaries, which disregard the operator's transient affective state and risk misaligned control interventions. To address this limitation, we propose an affect-aware shared autonomy teleoperation framework that dynamically modulates robotic assistance based on real-time operator state estimation. The system estimates operator affective states from synchronized facial video, cardiac signals, and bilateral arm kinematics, outputting a seven-state affective distribution and a three-category operational abstraction (neutral, productive, adverse). Affect-aware assistance is selectively triggered when the user is detected in a continuous adverse state, preserving task-positive engagement without unnecessary disruption. The empirical user study ($N = 30$) confirms that the proposed affective assistance increases the productive states by up to 39.7\% without compromising user agency. The collected dataset represents the first multimodal dataset that provides continuous visual, physiological, and operator's bilateral motion tracking of temporal affective state shifts during bimanual teleoperation. Our multimodal fusion model outperforms zero-shot baselines (Qwen, MiniCPM-V) in tracking temporal state dynamics. This real-world deployment offers a new human-centric framework that integrates visual, physiological, and motion tracking for physical human-robot interaction.
\end{abstract} 

\begin{figure*}[h]
  \centering
  \includegraphics[width=0.8\linewidth]{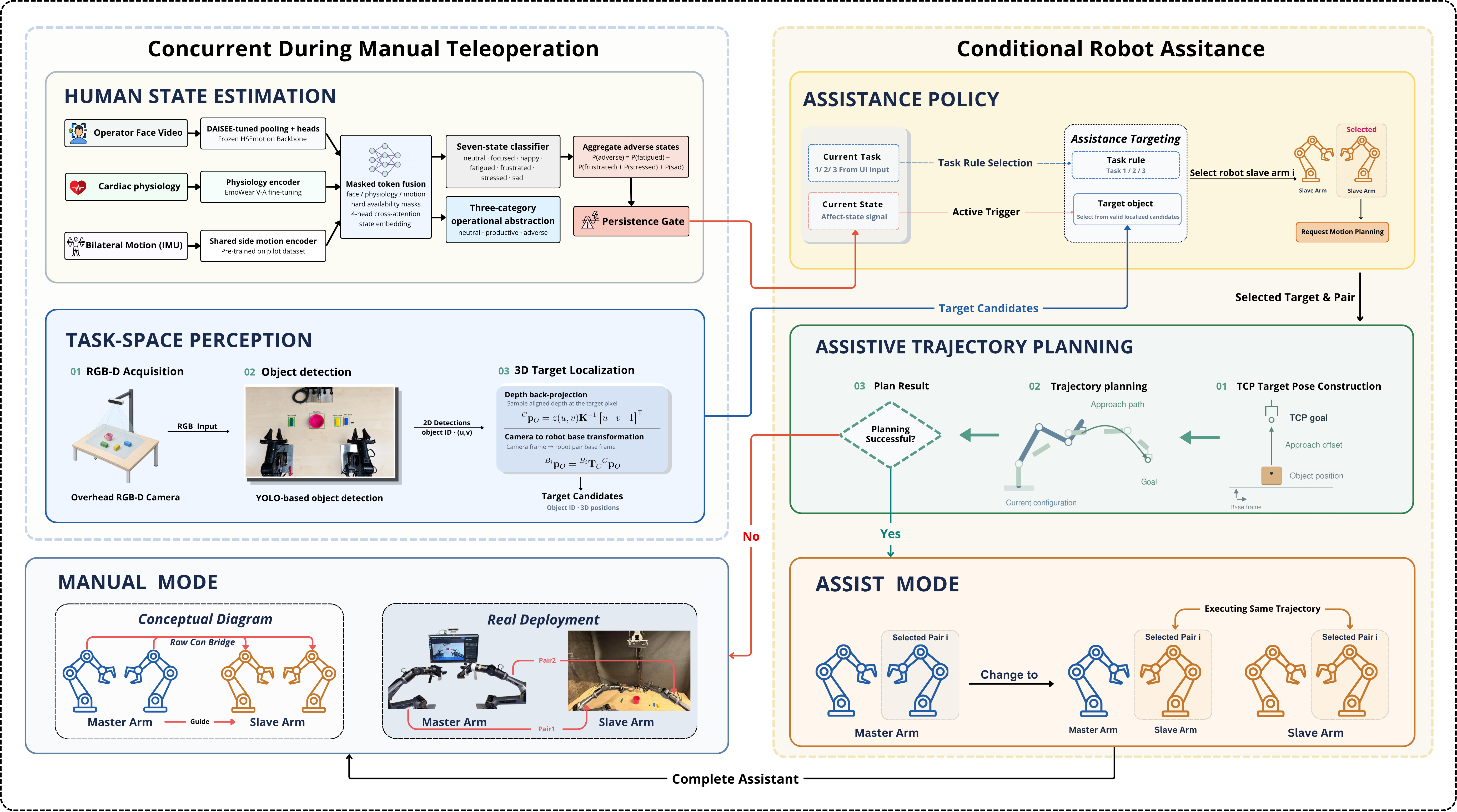}
  \vspace{-0.05in}
  \caption{Affect-aware Shared-Autonomy Framework}
  \vspace{-0.15in}
  \label{fig:framework}
\end{figure*}   

\section{INTRODUCTION}
\label{introduction}

Teleoperation bridges human cognitive versatility, spatial reasoning, and adaptability with the precision, physical payload capabilities, and reach of robotic systems. By maintaining supervisory or direct human-in-the-loop control, operators can perform high-dexterity manipulation tasks across hazardous or unconstructed environments, including space exploration \cite{SpaceApp2026}, robotic telesurgery \cite{RobotTelesurgery2025}, remote nuclear decommissioning \cite{NuclearApp2025}, and socially assistive robots \cite{AdaptiveBehaviorAcquisition2019}. In these teleoperated domains, task execution relies heavily on the seamless exchange of visual, kinematic, and haptic information between the operator and the master--slave interface. 

Despite advances in bilateral control and shared autonomy, remote manipulation remains a highly cognitive workload in practice. Existing assistance systems typically employ task-based triggers \cite{songTATIC2026}. These triggers respond to observable execution conditions, such as task errors. However, these approaches do not consider operator states, which may lead to assistance that is misaligned with the operator's actual needs \cite{brooks2019balanced}. 
While prior work has explored physical workload or cognitive demand in isolation, the complex interdependencies between task-induced stress, cognitive workload moderators, and human task performance remain insufficiently understood, particularly during shared control authority shifts~\cite{samStressWorkload2024}. Consequently, existing task- and state-based triggers lack clear guidelines for optimal intervention timing. Without insights into temporal affective dynamics across task execution, automated systems risk intervening abruptly during productive states. On the other hand, it might withhold critical aid during acute frustration and performance breakdown without temporal affective understanding.

To address this limitation, this study investigates the interaction between physiological stress markers, cognitive workload, and execution performance during high-precision bimanual teleoperation. Focused on task engagement, we define seven states and map them into a three-category operational abstraction (neutral contains \textit{neutral} state, productive includes \textit{focused} and \textit{happy} states, and adverse includes \textit{frustrated, fatigued, stressed} and \textit{sad} states). The operator states are used interchangeably with operator affective states in this study. Specifically, we investigate the following question: How do affective states fluctuate across task progress, and when is the ideal window for system intervention? Does an affect-aware intervention paradigm improve adaptation to human operators in terms of emotional transitions and perceived workload compared to static assistance?

To answer these research questions, we propose an affect-aware shared control teleoperation framework validated via an empirical user study ($N = 30$) across direct manual, geometry-driven, and affect-aware assistance modes. The primary contributions of this work are: 
\begin{itemize}
    \item \textbf{Affective Shared Autonomy Architecture:} We propose an affect-aware teleoperation framework that dynamically provides automated guidance based on real-time operator affect state estimation.
    \item \textbf{Multimodal Fused State Estimation:} We propose a multimodal human state estimation model that fuses facial, physiological, and motion modalities with a focus on task engagement and frustration. 
    \item \textbf{Multimodal Biometric and Behavioral Dataset\footnote{The completed dataset will be publicly available upon publication of this manuscript.}:} We introduce the first synchronized multimodal dataset for affect-aware bimanual teleoperation, covering facial-expression, physiological, and motion streams with temporal affect annotations in order to foster further research in affect-aware human-robot interaction designs.
    \item \textbf{Empirical Validation of Affect-Aware Assistance:} The results ($N=30$) show that affect-aware assistance increases operator productive state by up to 39.7\% relative to manual teleoperation while maintaining comparable task performance and perceived agency.
\end{itemize} 

\section{RELATED WORK}
\label{sec:related_work}
\subsection{Shared Autonomy}
Shared autonomy refers to robotic control frameworks in which control authority is distributed between the human operator and the autonomous system. In physical human-robot interaction (pHRI), shared-control systems commonly rely on intent detection, arbitration, and communication mechanisms to determine when and how assistance should be applied while preserving user agency \cite{loseyReviewIntentDetection2018}. A common formulation is continuous control blending, where the user's command and the robot's assistive policy are combined according to inferred intent or task context \cite{draganPolicyBlending2013,moustrisSharedControl2026}. In teleoperation and assistive manipulation, shared autonomy has also been studied as a way to automate selected components of a task while leaving meaningful control to the user. Vasile et al. proposed a vision-based shared-autonomy framework for prosthetic wrist control, where visual perception continuously assists wrist orientation during the approach-to-grasp phase while the user remains involved in phase switching and final grasp execution \cite{vasileContinuousWristControl2025}.

Recent work has further explored how assistive policies can be learned or adapted across repeated interactions. Tao et al. proposed an incremental learning framework for robot shared autonomy, first pretraining an assistive policy using simulated kinematic trajectories and then refining it through real-world user interactions \cite{taoIncrementalLearningRobot2025}. Related work on automatic mode switching has used task context and large language models to reduce the burden of manually selecting control modes during assistive teleoperation \cite{taoLAMS2025}. These studies show that shared autonomy is not only a low-level control problem, but also a question of when assistance should be invoked and how control authority should be allocated over time.

\subsection{Human Affective State Estimation}
Yue et al. emphasized that psychological safety is equally important as physical safety to achieve an optimal interaction in active pHRI scenarios. They also mentioned that it would be necessary to focus on those factors that can be measured with noninvasive sensors and data that can be easily obtained in future implementations \cite{huActivePhysicalHuman2022}. Physiological signals used in mental state detection are galvanic skin response and heart rate variability (HRV). A device is used to measure photoplethysmography (PPG) so that HRV and blood volume pulse can be calculated from PPG. Non-invasive sensors are an effective measurement tool for human-robot encounters \cite{guptaIndoorHRI2026}. The functionality of motions is also explored in mental state estimation \cite{wangFullBody2024}.

Given the high inter-subject variability in facial expressions, first-person retrospective review ensures reliable ground-truth annotations by reflecting the operator's true internal state. While multimodal emotion recognition datasets mainly rely on third-person annotations from external observers, first-person self-annotations by operators reviewing their own teleoperation recordings remain exceptionally rare~\cite{kim2025review}. {\em This gap is critical in pHRI}, where operators exhibit subtle, task-focused affect rather than discrete expressions, leading to frequent misclassification by standard models~\cite{koelstra2011deap, hong2026m}. Furthermore, existing teleoperation and shared-control studies often evaluate operator workload through post-trial questionnaires, capturing aggregated experiences rather than continuous within-task dynamics~\cite{panSharedControlLoadTrust2024}. Consequently, temporal affective state shifts across task execution remain largely unmapped. We address this gap by collecting task-grounded multimodal data, providing a basis for future operator‑state monitoring and robot assistance. 

\section{METHODOLOGY}
\label{sec:methodology}
The proposed affect-aware shared autonomy framework is illustrated in Fig.~\ref{fig:framework}. In manual control mode, the system continuously tracks task progress while simultaneously assessing operator affective states using a multimodal fusion model. When an assistance policy issues a request, the system selects the target object and robot pair, retrieves the corresponding localized target position, and attempts to generate an assistive trajectory. Upon completion of an assistance episode, control authority will be seamlessly handed back to the human operator, resetting the pipeline for continuous monitoring, planning, and execution.
\vspace{-0.8ex}
\subsection{Multimodal Human State Prediction}
A novel multimodal human state estimation model is developed for affect-aware shared autonomy. Since we focus on critical state shifts during task execution, especially moments when participants encounter operational difficulty and approach negative states, operator states are defined and grouped into three operational categories and seven affect states: 
\begin{itemize}
    \item \textbf{Neutral}: a baseline operational state without emotional activation, containing \textit{neutral} state.
    \item \textbf{Productive}: high-engagement states conducive to task success, including \textit{focused} and \textit{happy} states. 
    \item \textbf{Adverse}: negative states that degrade performance, including \textit{frustrated}, \textit{stressed}, \textit{fatigued}, and \textit{sad} states.
\end{itemize}
The estimator is designed to fuse operator-face video, cardiac physiology, and bilateral arm motion to produce probabilistic seven-state affect estimates and a three-category operational summary. The endpoint uses the Convolutional Neural Network (CNN) only in the frozen facial backbone and Multi-Layer Perceptrons (MLPs) for physiology and bilateral motion. Their embeddings are projected to common tokens and fused by masked multi-head attention. 

The aggregated latent multimodal vector $\mathbf{z}_t$ acts as a unified state embedding. To optimize predictive stability, our architecture decouples this representation into a \textit{hierarchical classification framework} via two parallel projection layers (heads). The seven-state probability distribution and three operational category probability distribution are defined as: 
\begin{equation}
\begin{aligned}
    \widehat{\mathbf p}_t &=\big[\widehat p_{\mathrm{neutral}}(t),\widehat p_{\mathrm{focused}}(t),\widehat p_{\mathrm{happy}}(t), \\
    & \widehat p_{\mathrm{fatigued}}(t),\widehat p_{\mathrm{frustrated}}(t),\widehat p_{\mathrm{stressed}}(t),\widehat p_{\mathrm{sad}}(t)\big]^{\mathsf T} \\
    \widehat{\mathbf r}_t &=\big[\widehat r_{\mathrm{neutral}}(t),\widehat r_{\mathrm{productive}}(t),\widehat r_{\mathrm{adverse}}(t) \big]^{\mathsf T}. 
\end{aligned}
\end{equation}
where $\sum_{c=1}^7 \widehat p_{t,c}=1$ and $\sum_{m=1}^3 \widehat r_{t,m}=1$. The $\widehat{\mathbf p}_t$ and $\widehat{\mathbf r}_t$ are \textit{estimated} by:
\begin{equation}
   \begin{aligned}
    \widehat{\mathbf{p}}_t &= \text{softmax}(\mathbf{W}_p \mathbf{z}_t + \mathbf{b}_p) \\
    \widehat{\mathbf{r}}_t &= \text{softmax}(\mathbf{W}_r \mathbf{z}_t + \mathbf{b}_r).
\end{aligned}
\label{eq:causal-output} 
\end{equation}
where $\mathbf{z}_t$ will be computed by Equation~(\ref{eq:masked-fusion}). $\hat{\mathbf{p}}_t \in \mathbb{R}^7$ denotes the predicted probability distributions over the seven affective states, and $\hat{\mathbf{r}}_t \in \mathbb{R}^3$ denotes the three operational categories, respectively. $W_p$ and $W_r$ are learned classifier weights. $b_p$ and $b_r$ denote the bias. Both distributions are estimated human-state variables for downstream decision support, not direct robot commands.

\subsubsection{Modality adaptation and encoding}
For each frame $I_{t,j}$ in the causal window $\mathbf{J}_t$, the frozen HSEmotion backbone $\mathbf{E}_F$ \cite{HSEmotion} returns $\mathbf h^F_{t,j}\in\mathbb R^{1280}$. DAiSEE \cite{gupta2022daisee} trains the pooling and ordinal heads for engagement, frustration, and confusion over frozen HSEmotion embeddings:
\begin{equation}
   \begin{aligned}
    e_{t,j} &= \mathbf{w}_a^{\mathsf{T}} \tanh(\mathbf{W}_a \mathbf{h}^F_{t,j} + \mathbf{b}_a) \\
    \alpha_{t,j} &= \frac{\exp(e_{t,j})}{\sum_{\ell \in \mathbf{J}_t} \exp(e_{t,\ell})} \\
    \widetilde{\mathbf{e}}^F_t &= \sum_{j \in \mathbf{J}_t} \alpha_{t,j} \mathbf{h}^F_{t,j} 
\label{eq:face-pooling}
\end{aligned} 
\end{equation}
where $e_{t,j}$ represents the raw scalar alignment, $\alpha_{t,j}$ is the normalized attention weight, and $\widetilde{\mathbf{e}}^F_t \in \mathbb{R}^{1280}$ is the pooled facial embedding. 
The downstream heads are optimized under threshold-based ordinal supervision for engagement, frustration, and confusion. The final facial token $\mathbf{e}^F_t \in \mathbb{R}^{1296}$ is constructed via vector concatenation ($\parallel$):
\begin{equation}
    \mathbf{e}^F_t = \left[ \widetilde{\mathbf{e}}^F_t \parallel \mathbf{c}_t \right]
\end{equation}
where $\mathbf{c}_t \in \mathbb{R}^{16}$ is an auxiliary context vector capturing the three DAiSEE class distributions, their expected ordinal intensities, and the base encoder's happiness probability.

In practical teleoperation environments, the physiological and kinematic data streams should tolerate intermittent sensor disconnects and lighting changes without breaking down. Rather than using sparse, hardcoded filtering rules, binary availability indicators are passed directly into the feature encoders. 
The physiological vector $\mathbf{x}^P_t$ comprises causal PPG moments, HRV, and respiration statistics, together with coverage, flatline, clipping, and RR-outlier descriptors. It maps the 16 masked features and their mask (32 inputs) through a 256-unit hidden layer to this 256-dimensional embedding. The physiological encoder maps raw physiological data \(\mathbf{x}_{t}^{p}\) conditioned on its sensor mask \(\mathbf{m}_{t}^{p}\):
\begin{equation}
    \mathbf{e}_t^p = \mathbf{E}_P(\mathbf{x}_t^p, \mathbf{m}_t^p), \quad \mathbf{e}_t^p \in\mathbb R^{256}
\end{equation}
$\mathbf{E}_P$ is trained with cross-entropy on EmoWear's four valence--arousal quadrants \cite{EmoWear2024} and then frozen for pilot-fusion training and deployment. Concurrently, the bilateral motion tracking network mirrors this paradigm across the left and right manipulator trajectories:
\begin{equation}
    \mathbf{e}_t^M = \mathbf{E}_M(\mathbf{x}_t^{s,\text{mo}}, \mathbf{m}_t^{s,\text{mo}}), \quad s \in \{L, R\}
\end{equation}
where \(\mathbf{m}_{t}^{s,\text{mo}}\) is the binary availability mask for motion features from side \(s\) (L: left, R: right) at time \(t\). This architecture ensures that if a specific stream fails or drops packages, its hidden states are structurally handled, preventing the propagation of corrupted values into the fusion block.
The seven-state labels are from the self-validated annotations in the pilot dataset, which includes a number of participants with unassisted manual control. EmoWear supervises physiology only. The pre-trained physiology encoder is fine-tuned on the pilot dataset. Bilateral motion is learned from pilot-state supervision within the fusion model. The shared weights preserve left-right symmetry and prevent a missing arm from being interpreted as a confident zero-valued signal. 

\subsubsection{Masked token fusion and training objective}
Each modality is linearly projected into a shared space $\mathbf{v}_{t,k} = \mathbf{P}_k\mathbf{e}_t^k + \mathbf{b}_k \in \mathbb{R}^{256}$, where $k \in \{F,P,M\}$. $F$ represents face modality, $P$ represents physiology modality, and $M$ represents motion modality. \(\mathbf{P}_{k}\) is a trainable linear projection matrix for modality \(k\). For each head $h \in \{1,\dots,H\}$, hard availability indicator $a_{t,k} \in \{0,1\}$, the $d_h$-dimensional slice $\mathbf{v}_{t,h,k}$ ($d_h=256/H$) is evaluated against a learnable query $\mathbf{q}_h \in \mathbb{R}^{d_h}$ via masked cross-attention:
\begin{equation}
\begin{aligned}
    \ell_{t,h,k} &= \frac{\mathbf{q}_h^{\mathsf{T}}\mathbf{v}_{t,h,k}}{\sqrt{d_h}} + \begin{cases}
0, & a_{t,k}=1,\\
-\infty, & a_{t,k}=0,
\end{cases} \\
    \beta_{t,h,k} &= \frac{\exp(\ell_{t,h,k})}{\sum_{j \in \{F,P,M\}} \exp(\ell_{t,h,j})} \\
    \mathbf{z}_t &= \mathbf{G}\left( \Big[ \sum_{k \in \{F,P,M\}} \beta_{t,h,k}\mathbf{v}_{t,h,k} \Big]_{h=1}^{H} \right) 
\label{eq:masked-fusion}
\end{aligned}   
\end{equation}
where $\ell_{t,h,k}$ is the scaled alignment score and $\beta_{t,h,k}$ is the normalized attention weight. $G$ denotes the post-attention projection that maps the concatenated multi-head output to $\mathbf z_t$.
The network is optimized end-to-end via a multi-task objective function $\mathbf{L}$ that balances participant, trial, and episodic distributions using a sample weight $\omega_t$ as
\begin{equation}
    \mathbf{L} = \sum_t \omega_t \left( \mathbf{L}_{\mathrm{r}} + \gamma \mathbf{L}_{\mathrm{p}} \right) \label{eq:pilot-loss}
\end{equation}
where \(\mathbf{L}_{\text{r}} = -\sum_c y_{t,c}^r \log \hat{\mathbf{r}}_{t,c}\) optimizes the three-category mapping, \(\mathbf{L}_{p} = -\sum_c y_{t,c}^s \log \hat{\mathbf{p}}_{t,c}\) supervises the fine-grained seven-state classifier, and $\gamma$ is a task-balancing hyperparameter experimentally determined. To enhance robustness, modality dropout randomly zeroes out one observed token during training.

\subsection{Affect-aware assistance design}
The affect-aware assistance path converts uncertain state estimates into a conservative,
reversible assistance request only after sustained adverse evidence. The proposed system first derives adverse probability from the seven-state output and provides live transmission to the shared-control coordinator:
\begin{equation}
\widehat p_{\mathrm{adverse}}(t)=\widehat p_{\mathrm{fatigued}}(t)+
\widehat p_{\mathrm{frustrated}}(t)+\widehat p_{\mathrm{stressed}}(t)+
\widehat p_{\mathrm{sad}}(t).
\label{eq:affect-adverse}
\end{equation}
The $\widehat p_{\mathrm{adverse}}(t)$ is an operational aggregation of adverse states at time $t$, not a separate stress classifier. With threshold $\theta \in [0,1]$, hold duration $H$ (seconds), maximum inter-prediction gap $g$ (seconds), and cooldown $C$ (seconds), an assistance episode begins when
\begin{equation}
    \text{Trigger}(t) = \left[ \min_{\tau \in [t-H, t]} p_{\text{adverse}}(\tau) > \theta \right] \cdot \left[ t - t_{\text{last}} \geq C \right]
    \label{eq:affect-trigger}
\end{equation}

Equation \eqref{eq:affect-trigger} requires continuous above-threshold evidence for one second. Any below-threshold prediction, non-monotonic timestamp, or gap exceeding $g$ resets the hold; $t_{\mathrm{last}}$ is the previous episode-trigger time. On triggering, the service assigns a unique episode identifier and requests fixed medium assistance. Assistance remains active until task completion, execution error, an explicit operator or safety stop, trial termination, or confirmed task exit. Declining adverse probability alone does not terminate an episode. 
We would like to note that it is a conservative affect-informed assistance prototype rather than a validated autonomous affect-response system. 

\subsection{Shared Autonomy Control Framework}
The platform comprises two master--slave arm pairs for bimanual teleoperation. During manual operation, the operator guides the master arms, whose motions are relayed to the corresponding slave arms. When assistance is activated, the selected pair switches to assistive trajectory control, while the other pair remains under manual teleoperation. For each pair, manual and assistive commands are mutually exclusive.

\paragraph{Object Detection and 3D Target Localization}
An overhead RGB-D camera provides color images and aligned depth observations of the task workspace. A YOLO-based detector identifies task-relevant objects in the color image. For each selected detection, the image coordinates and corresponding depth observations are used to estimate the three-dimensional position of the object in the camera frame. Let $C$ denote the camera frame, $B_i$ the base frame of robot pair $i$, and ${}^{C}\mathbf{p}_{O}=[x_C,y_C,z_C]^{\mathsf{T}}\in\mathbb{R}^{3}$ the detected object position in the camera frame. The object position is transformed into the corresponding robot base frame as
\begin{equation}
    \begin{bmatrix}
        {}^{B_i}\mathbf{p}_{O} \\
        1
    \end{bmatrix}
    =
    {}^{B_i}\mathbf{T}_{C}
    \begin{bmatrix}
        {}^{C}\mathbf{p}_{O} \\
        1
    \end{bmatrix}
    \label{eq:object_position_transform}
\end{equation}
where ${}^{B_i}\mathbf{T}_{C}$ denotes the calibrated transformation from the camera frame to the base frame of robot pair $i$. The perception module retains the latest valid localized position for each detected task-relevant object together with its timestamp. A cached target is used for assistance only if it remains valid and sufficiently recent. Otherwise, the assistance request is rejected before motion planning.

\paragraph{Assistive Trajectory Planning}
For each assistance episode, the assistance policy selects both the target object and the robot pair assigned to the action. The planner then retrieves the corresponding cached target position for the selected pair and applies a task-specific approach offset to define the desired tool center point (TCP) target position:
\begin{equation}
    {}^{B_i}\mathbf{p}^{*}_{\mathrm{TCP}}
    =
    {}^{B_i}\mathbf{p}_{O}
    +
    {}^{B_i}\mathbf{d}_{A}
    \label{eq:tcp_target_position}
\end{equation}
where ${}^{B_i}\mathbf{d}_{A}$ denotes the task-specific approach offset expressed in the base frame of robot pair $i$. The TCP frame accounts for the fixed geometric offset between the wrist link and the gripper interaction point. Given the current robot configuration, the planner computes an approach trajectory for the selected pair using the desired TCP position and a predefined top-down tool-orientation constraint. If no feasible trajectory is found, the assistance request is rejected without issuing trajectory commands. Otherwise, the planned trajectory is forwarded to the control-handover procedure.

\paragraph{Control Handover and Execution}
Before executing the planned trajectory, manual command transmission is suspended only for the selected pair, and the transition to trajectory control is verified. The master and slave arms of the selected pair then execute the planned motion synchronously, while the non-selected pair remains under manual teleoperation. The motion of the selected master arm provides the operator with a physical indication that assistance is active. Upon completion or interruption, trajectory execution is terminated, and the system initiates the return to manual teleoperation. All assisted experimental conditions use the same perception, planning, handover, and execution pipeline and differ only in the policy that determines when an assistance episode is triggered. 

\section{USER STUDY}
\label{sec:user_study}
To evaluate how different automated intervention paradigms affect operator affective states and task performance, we conduct a user study approved by our university's Human Research Ethics Committee. A total of 33 participants were recruited. Data from 3 participants is excluded prior to analysis due to technical system errors and participant availability constraints, resulting in a final dataset of $N = 30$ participants (22 males, 8 females) allocated equally across the three experimental groups ($10$ per group). The overall experimental protocol lasts approximately 90 minutes, including sensor personalized calibration, system training, trial execution, post-trial affective state self-annotation, and post-trial survey. 

\subsection{Experimental Setup}
\label{sec:experimental_setup}
\begin{figure}[h]
    \includegraphics[width=0.98\linewidth]{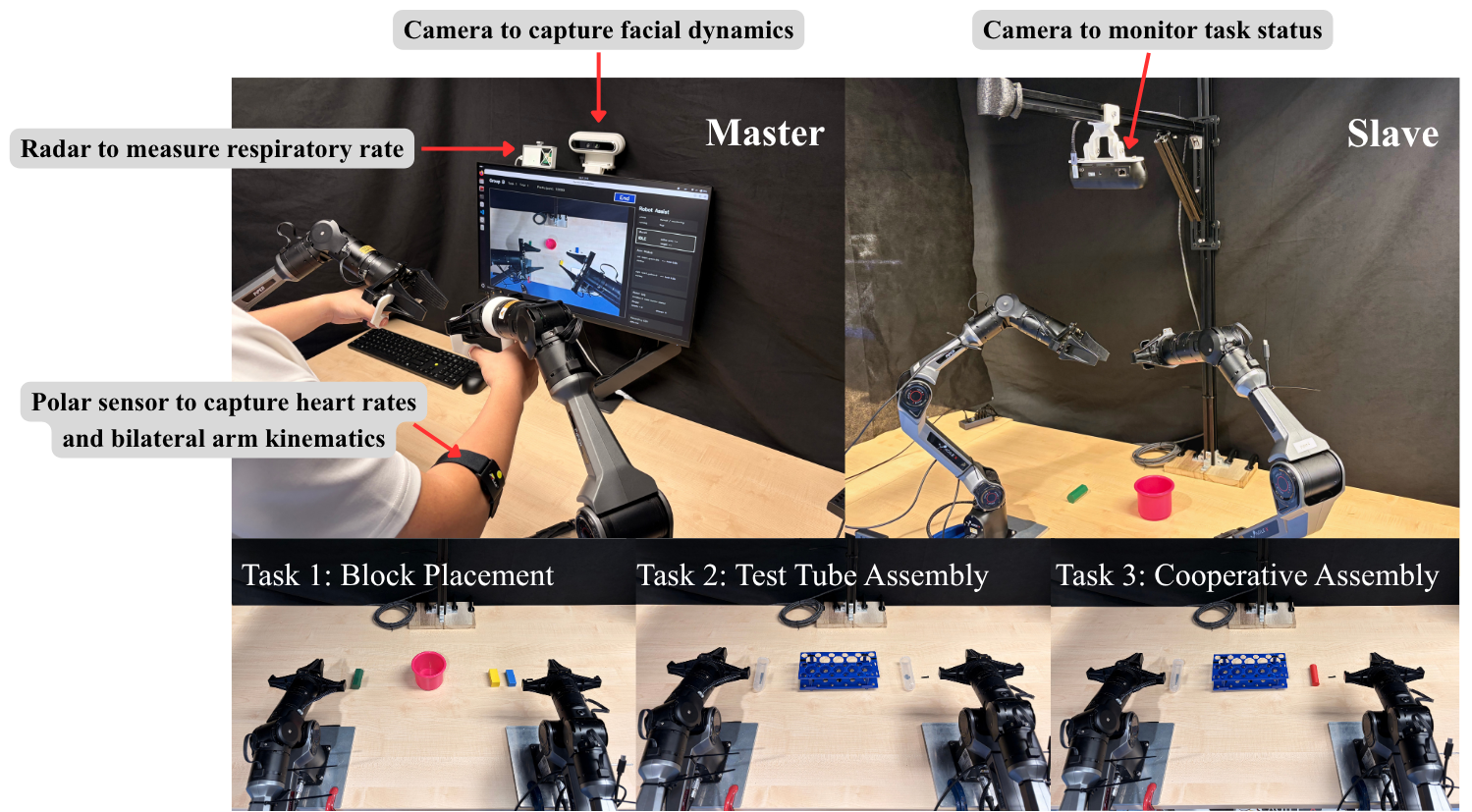}
    \vspace{-0.05in}
    \caption{Experimental Setup}
    \vspace{-0.05in}
    \label{fig:experimental_setup}
\end{figure}

As shown in Fig.~\ref{fig:experimental_setup}, the physical setup consists of a dual-arm master–slave teleoperation platform controlled through two master handles. The operator station is where the human operator stands to control the robots and monitor the task. During the experiment, the operator holds the end effectors of these 6-DOF robotic arms to input movements. In our human-in-the-loop framework, control authority can shift between the operator and the robotic assistance module. When assistance is activated, the handle provides tactile cues of robot intervention; when assisted control ends, authority returns to the operator. 
Participants complete all tasks using visual feedback only from a front-facing camera display with direct view of the blocked remote workspace.  
\vspace{-1ex}

Each participant completes three manipulation tasks, with three trials per task:
\begin{enumerate}
    \item Task 1: Pick up three colored blocks from the table and place them into a central pink cup.
    \item Task 2: Pick up the test tubes from the table and insert them into the holes in the central test tube rack.
    \item Task 3: Hold a test tube with one arm while using the other one to insert a red block into the tube opening.
\end{enumerate}  
Participants are randomly assigned to one of three groups and operate the dual-arm system in a controlled environment. Group A performs manual manipulation without assistance, Group B receives geometry-driven assistance, and Group C provides affect-aware assistance. Participants do not repeat tasks under all three conditions to avoid learning effects across conditions, as improvements after manual practice may be hard to distinguish from robot assistance. 

\subsection{Data Collection}
\label{sec:data_collection}
To capture operator affective dynamics during teleoperation, we develop a multimodal data acquisition framework combining visual and physiological sensing as well as a video-assisted retrospective self-annotation protocol. After neutral-state calibration, facial expressions are captured using an Intel RealSense D455f camera, and biometric data is collected using two Polar Verity Sense optical sensors positioned bilaterally on the operator’s forearms. Based on the open-source Polar BLE SDK, we develop a customized pipeline to stream real-time PPG, Accelerometer (ACC), Gyroscope (GYRO), and Magnetometer (MAG) data from each arm during each trial. Respiration rate was measured using a 60 GHz millimeter wave radar module \cite{HLK2024} positioned about 1 meter in front of the operator’s chest, enabling contactless sensing with minimal interference.

Upon each task's completion, participants review synchronized recordings and annotated the rule-based flagged timestamps of state shifts with their perceived internal states. By incorporating individual expression baselines and subjective self-appraisal, these participant-verified labels serve as evaluation ground truth. After completing all the tasks, participants complete a post-trial questionnaire on their overall experience across the three bimanual tasks, including mental and physical fatigue, cognitive stress, and task satisfaction. 

\begin{figure*}[h]
  \centering
  \includegraphics[width=0.85\linewidth]{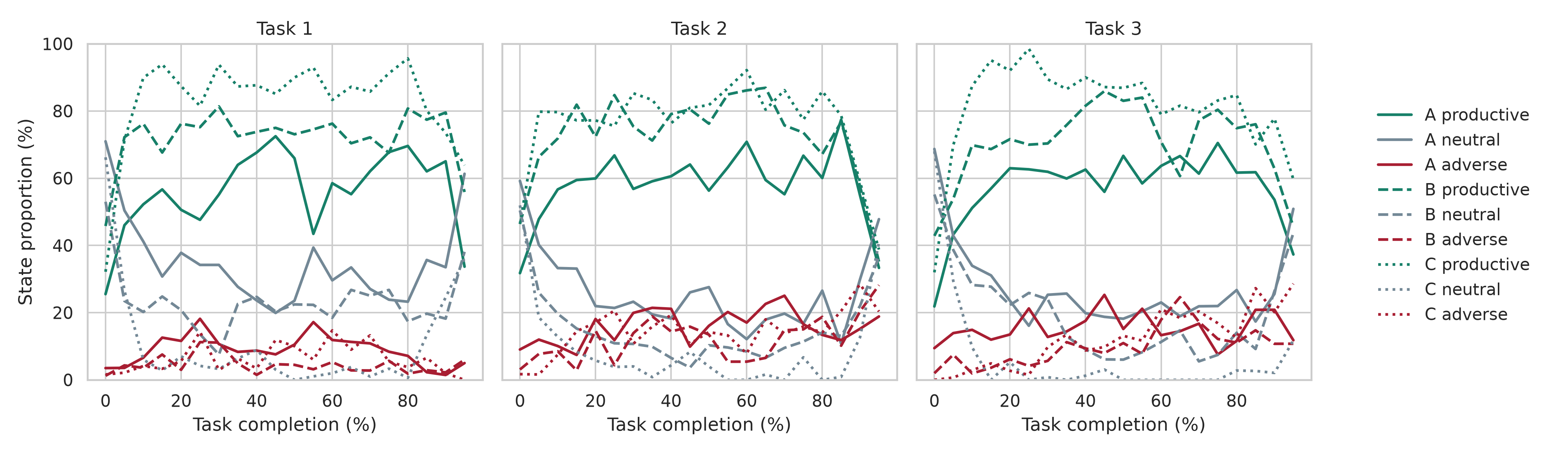}
  \vspace{-0.05in}
  \caption{Group Average Temporal State Proportion Comparison}
  \vspace{-0.05in}
  \label{fig:state_proportions}
\end{figure*}

\subsection{Evaluation Metrics}
\label{sec:evaluation_metrics}
The evaluation framework comprises two objective and subjective axes: temporal affective state proportions and subjective workload and usability surveys.

To investigate affective state transitions during task progression independently of raw execution speed, event annotations are mapped to normalized task progress bins rather than clock time. For annotation event $e$ occurring at timestamp $T_{i,t,r,e}$ (in seconds), the normalized task completion percentage $P_{i,t,r,e}$ is defined as $P_{i,t,r,e} = \min\left(100, \max\left(0, \frac{100 \cdot T_{i,t,r,e}}{V_{i,t,r}}\right)\right)$.

Completion percentages are discretized into 5\% task-progress bins $b \in \{0, 5, 10, \dots, 95\}$ via $b = 5 \cdot \left\lfloor P_{i,t,r,e} / 5 \right\rfloor$. 
Let $N_{i,g,t,r,b,s}$ denote the number of annotations for participant $i$ in group $g$, task $t$, trial $r$, and progress bin $b$ categorized under state $s \in \{\text{Neutral}, \text{Productive}, \text{Adverse}\}$. The relative state proportion $S_{i,g,t,r,b,s}$ (expressed as a percentage) is computed over the total bin annotations $N_{i,g,t,r,b,\text{total}} = \sum_{s} N_{i,g,t,r,b,s}$ as 
$S_{i,g,t,r,b,s} = 100 \times \frac{N_{i,g,t,r,b,s}}{N_{i,g,t,r,b,\text{total}}}$, where $\sum_{s} S_{i,g,t,r,b,s} = 100\%$ within every non-empty bin $b$. Group-level trajectory curves $\bar{S}_{g,t,b,s}$ are obtained by computing the mean state proportion across all trial instances within group $g$ for task $t$ and bin $b$. High localized proportions reflect consensus in human self-annotations regarding the operator's predominant state at specific phases of task progression. 

Subjective operator experience was assessed post-trial using a modified NASA-TLX \cite{2006nasa-tlx} survey structure evaluating four primary dimensions: mental and perceptual workload, physical fatigue, adverse affect, and task satisfaction. All survey dimensions were recorded on a standardized 11-point Likert scale ranging from $0$ (\textit{Strongly Disagree / No burden}) to $5$ (\textit{Neutral / Moderate burden}) and $10$ (\textit{Strongly Agree / Extreme burden}). Subjective responses per group are reported using descriptive statistics, including mean, standard deviation ($\text{SD}$), median, and interquartile range ($\text{IQR}$).

\section{RESULT ANALYSIS}
\label{sec:result_analysis}

\subsection{Temporal Affective State Proportions}
Operator annotation shows that productive states were the dominant annotation category in all groups (see Table~\ref{tab:state_proportions}). By task, Task 2 (Test Tube Rack Assembly) has the lowest mean completion fractions (A:0.53, B:0.57, C:0.40). This task difficulty is further evidenced by group mean average time (A:155.17s, B:133.66s, C:139.53s). Adverse proportions are comparatively low in all groups, while Group B and Group C were lower than unassisted manual control (Group A). In addition, affect-aware teleoperation (Group C) achieves the highest overall \textit{Productive} state ratio (79.2\%), which improves by up to $(\frac{79.2}{56.7}-1)*100\% =39.7\%$ compared to other groups. It demonstrates that affect-aware assistance enhances positive operator engagement during complex manipulation tasks. 

Looking into the state proportion shifts alongside the task completion percentages, one can see how affective dynamics fluctuate during sequential task phases (Fig.~\ref{fig:state_proportions}). At the beginning of each task, the neutral state is the dominant one. As the task progresses, the participants enter a focused state, which aligns with the productive category. For the challenging task 2, \textit{Adverse} state proportions are observed in the localized peaks. These spikes correlate directly with high-precision sub-goal steps, specifically the sequential fine-alignment and grasping phase for individual test tubes. Approaching the end of the task, a significant drop in \textit{Productive} proportions was observed across all groups, reflecting psychological release and cognitive offloading upon achieving task goals. 

A distinct temporal pattern in Groups B and C emerged regarding \textit{Adverse} state progression. Under manual control (Group A), negative affect exhibited transient cyclical fluctuations but returned to baseline. Conversely, fixed assistance (Group B) displayed a slight upward trend in \textit{Adverse} states near trial completion. We infer that this behavior is associated with perceptual uncertainty about the timing of automated interventions. Despite our design of pop-up reminders in the UI, the participants' intense focus on the task camera led them to overlook the assistance messages. This also led to minor control conflicts when unannounced assistance triggering occurred. 
Furthermore, affect-aware teleoperation (Group C) demonstrates a more bimodal state distribution, exhibiting a lower \textit{Neutral} proportion compared to Groups A and B. Consequently, while Group C maximizes overall \textit{Productive} engagement, it also registers a marginal increase in transient \textit{Adverse} states relative to Group B. Overall, these temporal trends confirm that automated shared autonomy boosts sustained operator focus and productivity during bimanual manipulation.
\vspace{-0.8ex}
\begin{table}[htbp]
    \centering
    \caption{Mean annotated state proportions by group}
    \vspace{-0.05in}
    \label{tab:state_proportions}
    \begin{tabular}{lccc}
    \toprule
    Group & Productive (\%) & Neutral (\%) & Adverse (\%) \\
    \midrule
    A & 56.7 & 29.9 & 13.4 \\
    B & 71.4 & 19.6 & \textbf{9.0} \\
    C & \textbf{79.2} & \textbf{9.9} & 11.0 \\
    \bottomrule
    \end{tabular}
    \vspace{-0.1in}
\end{table}
\vspace{-0.8ex}
\subsection{Post-trial Survey}
\begin{figure}[H]
  \centering
  \includegraphics[width=3.6in]{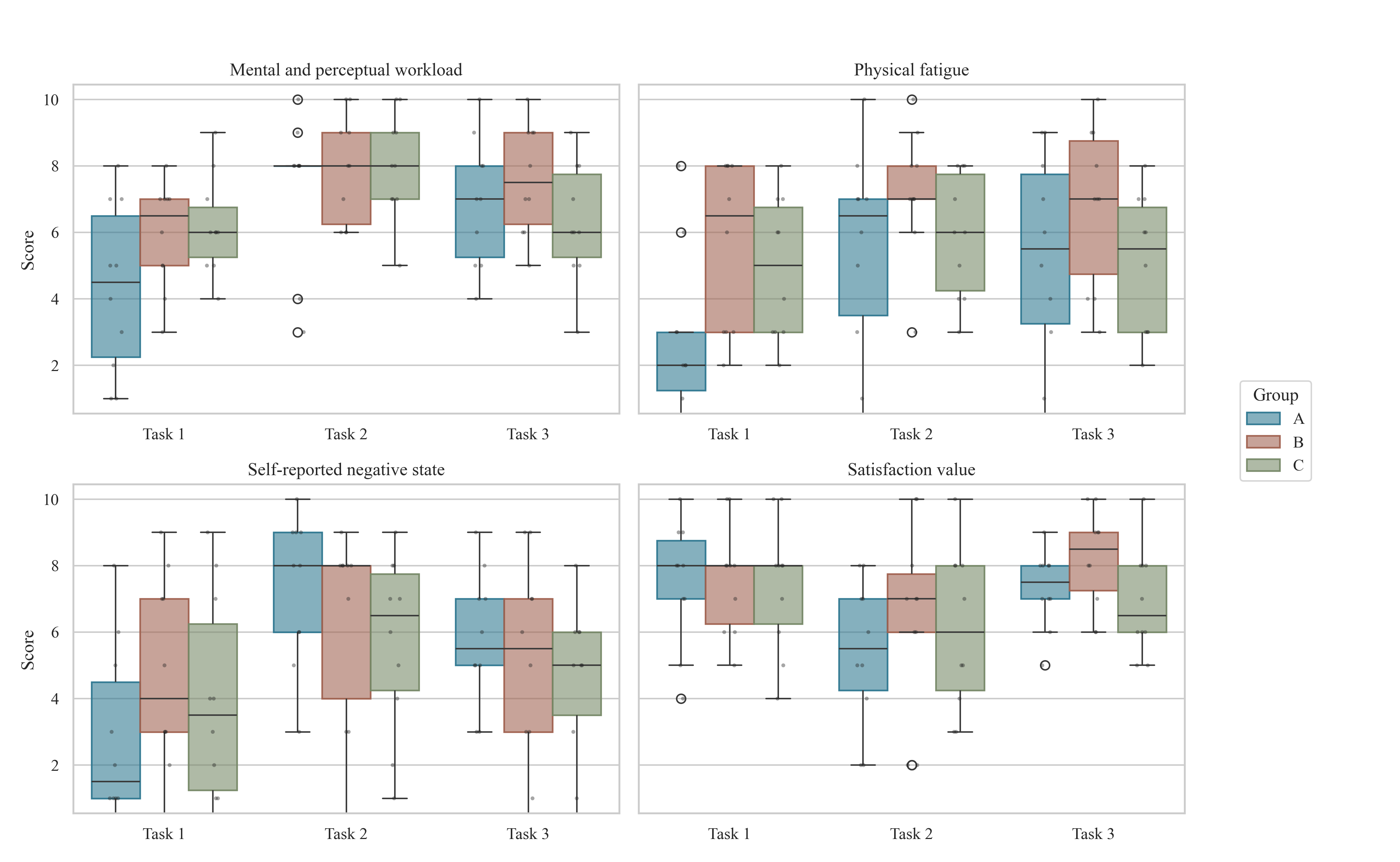}
  \vspace{-0.1in}
  \caption{Group Subjective Workload and Utility Comparison}
  \vspace{-0.1in}
  \label{fig:post_trial_survey}
\end{figure}

The post-trial questionnaire reveals distinct task-dependent workload variations and clear subjective differences across assistance conditions (Fig.~\ref{fig:post_trial_survey}). 
\paragraph{Perceived Cognitive and Physical Workload} 
Descriptively, mental workload increases from Task 1 to Task 2 in all groups, with mean scores changing from 4.30 to 7.40 in Group A, 5.90 to 7.90 in Group B, and 6.20 to 8.00 in Group C. Workload then remains high or declines modestly in Task 3. Notably, unassisted manual teleoperation (Group A) registers lower mean mental and physical workload scores during initial task execution compared to both assisted paradigms (Groups B and C). Furthermore, the median value of physical fatigue is elevated during Task 1 for participants in the assisted groups. This initial spike in workload is due to the cognitive and physical requirements of learning and adapting to the dynamics of autonomous intervention. Control negotiation with an active assistant can lead to temporary cognitive friction and motor co-contraction if human operators are not aware of the shifts in shared control authority until the user adapts. 
\paragraph{Adverse Affect and Overall Task Satisfaction} 
Evaluations of subjective adverse affect demonstrate the clear benefits of state-adaptive assistance over task progress. In Task 1, Group A reports the lowest negative state, whereas static geometry-driven assistance (Group B) induces the highest adverse affect. State-adaptive control (Group C) has a lower negative state than Group B during Task 1, demonstrating better initial alignment with operator intent. As task difficulty increased in Tasks 2 and 3, Group C achieved the lowest subjective adverse state among all three groups. This finding confirms that dynamically tailoring system intervention to detected operator affective states effectively mitigates task-induced frustration during complex manipulation. 

The analysis of control and dominance ratings is crucial, as it reveals that automated affective guidance did not undermine the operator's subjective sense of agency. Group~C had the highest composite dominance score (7.47), followed by Groups~B (6.70) and A (6.07); the same ordering appeared for task mastery (C:8.40, B:7.10, and A:6.10) and overall control (C:7.40, B:6.50, and A:5.80). Direct steering responsiveness was comparable across groups (6.30--6.60).

Despite achieving superior state mitigation and sense of control, overall task satisfaction in Group C remained slightly lower than in Group B. We attribute this satisfaction discrepancy to the dynamic adaptation curve inherent to affective shared autonomy. Because personalized affective models require iterative calibration to converge on individual user preferences, the time-constrained laboratory trial protocol limited the system's ability to fully co-adapt with each participant. Negotiating variable assistance policies under strict temporal constraints tended to introduce mild unpredictability during early task execution, slightly tempering overall satisfaction despite high reported agency and mastery.

\subsection{State estimation methods comparison}
\label{sec:state_estimation_comparison} 
State-estimation methods are evaluated on 5,107 annotated windows and features in Group~C. Predictions were matched independently to annotations from the same participant, group, task, and trial when their timestamps differed by at most 3~s. The \textit{sad} state is grouped with \textit{stressed} state in this baseline comparison. In addition, zero-shot methods are restricted to three-choice exploratory baselines because they cannot predict focused, happy, or frustrated. Accuracy reflects performance on prevalent states, while Weighted-F1 captures performance on the dominant temporal state distribution. Metrics are calculated from the confusion matrix. This comparison audits matched deployment-path predictions against annotations.
\begin{table}[h]
\centering
\caption{State prediction accuracy comparison on Group~C}
\vspace{-0.05in}
\label{tab:groupC-state-comparison}
\begin{tabular}{lcc}
\toprule
Method & Accuracy  & \makecell{Weighted-\\F1} \\
\midrule
Proposed fusion model & \textbf{0.259} & \textbf{0.301} \\
HSEmotion (facial-only) \cite{HSEmotion} & 0.142  & 0.050 \\
NormWear (physiological-only) \cite{NormWear} & 0.077  & 0.013 \\
Qwen2.5-Omni \cite{Qwen2.5-Omni} & 0.098 & 0.024 \\
MiniCPM-V 4.5 \cite{MiniCPM-V} & 0.101 & 0.026 \\
\bottomrule
\end{tabular}
\end{table}
\begin{figure}
    \centering
    \includegraphics[width=0.9\linewidth]{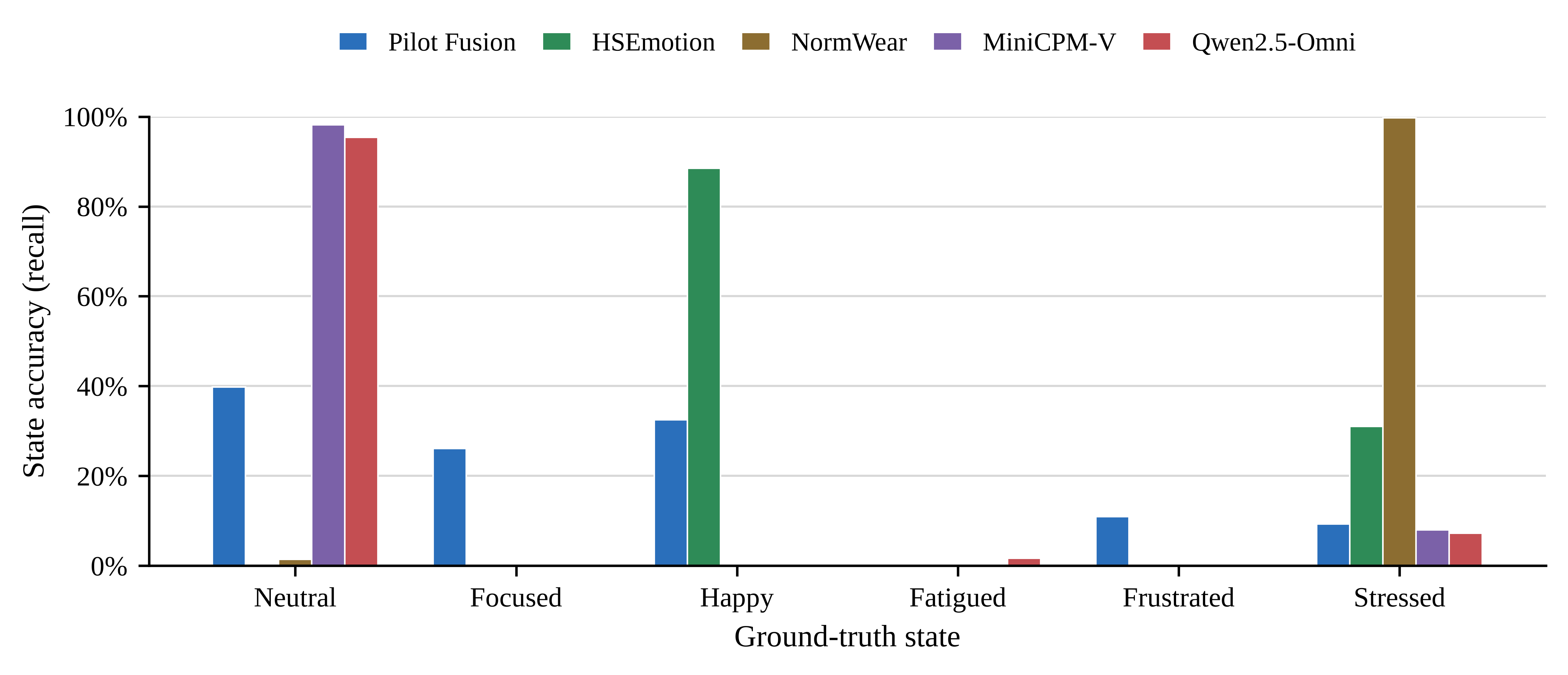}
    \vspace{-0.08in}
    \caption{Affective State Prediction Accuracy Comparison}
    \vspace{-0.2in}
    \label{fig:state_accuracy}
\end{figure}

Table~\ref{tab:groupC-state-comparison} shows that the proposed fusion model achieved the relatively strongest performance across all reported metrics. Its improvement in accuracy and Weighted-F1 indicates better class-balanced discrimination than the baseline methods. Nevertheless, the absolute scores remain modest, highlighting the difficulty of seven-state affect prediction under class imbalance and heterogeneous multimodal observations. According to Fig.~\ref{fig:state_accuracy}, the zero-shot baselines' predictions are imbalanced: NormWear is only excellent in capturing stressed signals (99\% accuracy), and MiniCPM-V has the highest accuracy in recognizing neutral states. However, the proportion of stressed and neutral states is small (Table~\ref{tab:state_proportions}), resulting in the low weighted F1 of baselines. Among the zero-shot baselines, HSEmotion achieved the highest accuracy (0.142) and the highest weighted F1 (0.050).

\subsection{Discussion}
\label{sec:discussion} 

To the best of the authors' knowledge, the dataset collected in this study represents the first multimodal dataset providing continuous visual, physiological, and motion tracking of temporal affective state shifts during bimanual teleoperation. The modest overall accuracy of real-time affect prediction stems from class-imbalanced annotation distribution, inter-subject variability, and decoupling between external expressiveness and subjective internal state. Operators spend the majority of task duration in \textit{focused} state, while adverse states (\textit{frustrated}, \textit{stressed}, \textit{fatigued}, and \textit{sad}) occur sparsely during transient manipulation bottlenecks. Individual facial expressiveness varies significantly. For example, some operators display facial tension during productive concentration, whereas others maintain a smiling expression during internal confusion. These challenges underscore the need for subject-calibrated, dynamic affective models in future shared autonomy systems. 

\section{CONCLUSION}
\label{sec:conclusion}
In this work, we develop a novel affect-aware shared control teleoperation framework that dynamically modulates robotic assistance based on real-time operator affective state estimation. To predict the affective state in real-time, we propose a multimodal model fused from facial video, cardiac signals, and bilateral arm kinematics. It outperforms zero-shot baseline models in affective state prediction. Through a user study ($N = 30$), we demonstrate that incorporating probabilistic affective modeling increases the proportion of operator productive states up to 39.7\% and preserves user agency compared to static assistance baselines. The dataset in this study presents the first multimodal dataset. It enables the evaluation of real-time affective state estimation algorithms or decision boundaries for optimal shared autonomy intervention timing, fostering further research in affect-aware human-robot interaction designs.






\bibliographystyle{IEEEtranBST/IEEEtran} 
\bibliography{References}

\begin{thebibliography}{10}
\providecommand{\url}[1]{#1}
\csname url@rmstyle\endcsname
\providecommand{\newblock}{\relax}
\providecommand{\bibinfo}[2]{#2}
\providecommand\BIBentrySTDinterwordspacing{\spaceskip=0pt\relax}
\providecommand\BIBentryALTinterwordstretchfactor{4}
\providecommand\BIBentryALTinterwordspacing{\spaceskip=\fontdimen2\font plus
\BIBentryALTinterwordstretchfactor\fontdimen3\font minus \fontdimen4\font\relax}
\providecommand\BIBforeignlanguage[2]{{%
\expandafter\ifx\csname l@#1\endcsname\relax
\typeout{** WARNING: IEEEtran.bst: No hyphenation pattern has been}%
\typeout{** loaded for the language `#1'. Using the pattern for}%
\typeout{** the default language instead.}%
\else
\language=\csname l@#1\endcsname
\fi
#2}}

\bibitem{SpaceApp2026}
L.~Mayershofer, F.~S. Lay, N.~Batti, S.~Brinkman, J.~Butterfa{\ss}, T.~Ehlert, E.~D. Exter, W.~Friedl, T.~Gumpert, A.~K{\"o}pken, X.~Luo, A.~N. Manaparampil, A.~Raffin, A.~Schmidt, F.~Schmidt, L.~Sch{\"u}rmann, D.~Seidel, R.~Luz, A.~S. Bauer, P.~Schmaus, D.~Leidner, T.~Kr{\"u}ger, and N.~Y. Lii, ``{Toward improving task-level commanding in space robotics teleoperation through shared mental models},'' in \emph{2026 IEEE Aerospace Conference}, 2026, pp. 1--15.

\bibitem{RobotTelesurgery2025}
S.~Schmidgall, J.~D. Opfermann, J.~W. Kim, and A.~Krieger, ``{Will your next surgeon be a robot? Autonomy and AI in robotic surgery},'' \emph{Science Robotics}, vol.~10, no. 104, p. eadt0187, 2025.

\bibitem{NuclearApp2025}
A.~Kenan, P.~Bremner, and M.~Giuliani, ``{Robot teleoperation design requirements from end users in nuclear facilities},'' in \emph{2025 IEEE/RSJ International Conference on Intelligent Robots and Systems (IROS)}, 2025, pp. 6749--6756.

\bibitem{AdaptiveBehaviorAcquisition2019}
M.~Hirokawa, A.~Funahashi, Y.~Itoh, and K.~Suzuki, ``{Adaptive behavior acquisition of a robot based on affective feedback and improvised teleoperation},'' \emph{IEEE Transactions on Cognitive and Developmental Systems}, vol.~11, no.~3, pp. 405--413, Sept. 2019.

\bibitem{songTATIC2026}
J.~Song, X.~Liang, and M.~Zheng, ``{TATIC}: Task-aware temporal learning for human intent inference from physical corrections in human-robot collaboration,'' \emph{arXiv preprint arXiv:2603.11077}, 2026.

\bibitem{brooks2019balanced}
C.~Brooks and D.~Szafir, ``{Balanced information gathering and goal-oriented actions in shared autonomy},'' in \emph{2019 14th ACM/IEEE International Conference on Human-Robot Interaction (HRI)}.\hskip 1em plus 0.5em minus 0.4em\relax IEEE, 2019, pp. 85--94.

\bibitem{samStressWorkload2024}
Y.~T. Sam, E.~{Hedlund-Botti}, M.~Natarajan, J.~Heard, and M.~Gombolay, ``{The impact of stress and workload on human performance in robot teleoperation tasks},'' \emph{IEEE Transactions on Robotics}, vol.~40, pp. 4725--4744, 2024.

\bibitem{loseyReviewIntentDetection2018}
D.~P. Losey, C.~G. McDonald, E.~Battaglia, and M.~K. O'Malley, ``{A review of intent detection, arbitration, and communication aspects of shared control for physical human--robot interaction},'' \emph{Applied Mechanics Reviews}, vol.~70, no.~1, p. 010804, Jan. 2018.

\bibitem{draganPolicyBlending2013}
A.~D. Dragan and S.~S. Srinivasa, ``{A policy-blending formalism for shared control},'' \emph{The International Journal of Robotics Research}, vol.~32, no.~7, pp. 790--805, 2013.

\bibitem{moustrisSharedControl2026}
G.~Moustris and C.~Tzafestas, ``{A shared-control framework for a human-robot front-following behaviour in unknown dynamic environments},'' \emph{International Journal of Social Robotics}, vol.~18, no.~2, p.~31, Feb. 2026.

\bibitem{vasileContinuousWristControl2025}
F.~Vasile, E.~Maiettini, G.~Pasquale, N.~Boccardo, and L.~Natale, ``{Continuous wrist control on the Hannes prosthesis: A vision-based shared autonomy framework},'' in \emph{2025 {{IEEE International Conference}} on {{Robotics}} and {{Automation}} ({{ICRA}})}.\hskip 1em plus 0.5em minus 0.4em\relax IEEE, May 2025, pp. 15\,107--15\,113.

\bibitem{taoIncrementalLearningRobot2025}
Y.~Tao, G.~Qiao, D.~Ding, and Z.~Erickson, ``Incremental learning for robot shared autonomy,'' \emph{arXiv preprint arXiv:2410.06315}, 2025.

\bibitem{taoLAMS2025}
Y.~Tao, J.~Yang, D.~Ding, and Z.~Erickson, ``{LAMS: LLM-driven automatic mode switching for assistive teleoperation},'' in \emph{2025 20th ACM/IEEE International Conference on Human-Robot Interaction (HRI)}.\hskip 1em plus 0.5em minus 0.4em\relax IEEE, 2025, pp. 242--251.

\bibitem{huActivePhysicalHuman2022}
Y.~Hu, N.~Abe, M.~Benallegue, N.~Yamanobe, G.~Venture, and E.~Yoshida, ``{Toward active physical human--robot interaction: Quantifying the human state during interactions},'' \emph{IEEE Transactions on Human-Machine Systems}, vol.~52, no.~3, pp. 367--378, June 2022.

\bibitem{guptaIndoorHRI2026}
R.~Gupta, E.~Norman, H.~Shin, Z.~Deng, M.~Esteva, N.~Lu, K.~K. Stephens, and L.~Sentis, ``{Indoor human--mobile robot encounters: A transdisciplinary study on perceived safety},'' \emph{ACM Transactions on Human-Robot Interaction}, vol.~15, no.~4, pp. 1--41, July 2026.

\bibitem{wangFullBody2024}
T.~Wang, S.~Liu, F.~He, W.~Dai, M.~Du, Y.~Ke, and D.~Ming, ``Emotion recognition from full-body motion using multiscale spatio-temporal network,'' \emph{IEEE Transactions on Affective Computing}, vol.~15, no.~3, pp. 898--912, July 2024.

\bibitem{kim2025review}
H.~Kim, Y.~Bian, and E.~G. Krumhuber, ``{A review of 25 spontaneous and dynamic facial expression databases of basic emotions},'' \emph{Affective Science}, vol.~6, no.~2, pp. 380--394, 2025.

\bibitem{koelstra2011deap}
S.~Koelstra, C.~Muhl, M.~Soleymani, J.-S. Lee, A.~Yazdani, T.~Ebrahimi, T.~Pun, A.~Nijholt, and I.~Patras, ``{DEAP: A database for emotion analysis; using physiological signals},'' \emph{IEEE Transactions on Affective Computing}, vol.~3, no.~1, pp. 18--31, 2011.

\bibitem{hong2026m}
S.~C. Hong, D.~Kuli{\'c}, and L.~Tian, ``{I'm not mad, just focused: Understanding human emotions in human-robot collaboration},'' \emph{IEEE Robotics and Automation Letters}, 2026.

\bibitem{panSharedControlLoadTrust2024}
J.~Pan, J.~Eden, D.~Oetomo, and W.~Johal, ``Effects of shared control on cognitive load and trust in teleoperated trajectory tracking,'' \emph{IEEE Robotics and Automation Letters}, vol.~9, no.~6, pp. 5863--5870, 2024.

\bibitem{HSEmotion}
A.~Savchenko, ``Facial expression recognition with adaptive frame rate based on multiple testing correction,'' in \emph{Proceedings of the 40th International Conference on Machine Learning (ICML)}, ser. Proceedings of Machine Learning Research, vol. 202.\hskip 1em plus 0.5em minus 0.4em\relax PMLR, 2023, pp. 30\,119--30\,129.

\bibitem{gupta2022daisee}
A.~Gupta, A.~D'Cunha, K.~Awasthi, and V.~Balasubramanian, ``{DAiSEE}: Towards user engagement recognition in the wild,'' \emph{arXiv preprint arXiv:1609.01885}, 2022.

\bibitem{EmoWear2024}
M.~H. Rahmani, M.~Symons, O.~Sobhani, R.~Berkvens, and M.~Weyn, ``{EmoWear: Wearable physiological and motion dataset for emotion recognition and context awareness},'' \emph{Scientific Data}, vol.~11, no.~1, p. 648, June 2024.

\bibitem{HLK2024}
\emph{HLK-LD6002 Respiration and Heart Rate Detection Radar Module Specification V1.1}, Hi-Link Technology, 2024, available: \url{https://www.hlktech.net/index.php?cate=cate-63fdaeb91b37b&id=download-center}.

\bibitem{2006nasa-tlx}
S.~G. Hart, ``{NASA-task load index (NASA-TLX); 20 years later},'' in \emph{Proceedings of the Human Factors and Ergonomics Society Annual Meeting}, vol.~50, no.~9, 2006, pp. 904--908.

\bibitem{NormWear}
Y.~Luo, Y.~Chen, A.~Salekin, and T.~Rahman, ``Toward foundation model for multivariate wearable sensing of physiological signals,'' \emph{ACM Transactions on Computing for Healthcare}, 2026.

\bibitem{Qwen2.5-Omni}
J.~Xu, Z.~Guo, J.~He, H.~Hu, T.~He, S.~Bai, K.~Chen, J.~Wang, Y.~Fan, K.~Dang, B.~Zhang, X.~Wang, Y.~Chu, and J.~Lin, ``{Qwen2.5-Omni} technical report,'' \emph{arXiv preprint arXiv:2503.20215}, 2025.

\bibitem{MiniCPM-V}
T.~Yu, Z.~Wang, C.~Wang, F.~Huang, W.~Ma, Z.~He, T.~Cai, W.~Chen, Y.~Huang, Y.~Zhao, \emph{et~al.}, ``{MiniCPM-V} 4.5: Cooking efficient {MLLMs} via architecture, data, and training recipe,'' \emph{arXiv preprint arXiv:2509.18154}, 2025.

\end{thebibliography}

\end{document}